# Rail break and derailment prediction using Probabilistic Graphical Modelling

R.M.C Taylor
*Transnet Freight Rail, Johannesburg, South Africa*

J.A. du Preez
*Stellenbosch University, Cape Town, South Africa*

ABSTRACT: Rail breaks are one of the most common causes of derailments internationally. This is no different for the South African Iron Ore line. Many rail breaks occur as a heavy-haul train passes over a crack, large defect or defective weld. In such cases, it is usually too late for the train to slow down in time to prevent a derailment. Knowing the risk of a rail break occurring associated with a train passing over a section of rail allows for better implementation of maintenance initiatives and mitigating measures. In this paper the Ore Line's specific challenges are discussed and the currently available data that can be used to create a rail break risk prediction model is reviewed. The development of a basic rail break risk prediction model for the Ore Line is then presented. Finally the insight gained from the model is demonstrated by means of discussing various scenarios of various rail break risk. In future work, we are planning on extending this basic model to allow input from live monitoring systems such as the ultrasonic broken rail detection system.

## 1 INTRODUCTION

One of the main goals of a rail freight company is to safely transport as much freight as the infrastructure and rolling stock can support. With advancement in infrastructure such as increased capacity of the overhead electrical supply, higher throughput signalling systems, and more advanced rolling stock (locomotives and wagons), the pressure on companies to move more freight reliably is increasing (Schafer 2009). One of the main factors that limit companies from achieving their freight targets is derailment of freight trains.

Derailments do not only cause freight companies to lose the freight in the derailed wagons, but they also cause major infrastructure damage that can take weeks to repair. The cost of repairing the infrastructure and rolling stock, as well as the losses from tonnages that cannot be moved until the infrastructure is back in place, can amount to billions of Rands for a severe derailment (such as one over a bridge) and tens of millions for a less serious derailment. Furthermore, derailments pose a severe safety risk to people working close to the track as well as the train driver.

Internationally, a rail break is often the primary cause of a derailment (Schafer 2009, Liu et al. 2014, Duvel 2015, Liu et al. 2012, Dick et al. 2001). Rail breaks are therefore known to be one of the most costly and dangerous rail infrastructure failures (Liu et al. 2014). However, not all rail breaks result in derailments (Jeong 2003).

Freight companies implement maintenance interventions to mitigate the risk of rail breaks and derailments (Liu et al. 2014), but numerous derailments still occur annually on railways all over the world. To drastically reduce the number of annual rail breaks the risk of a rail break occurring needs to be predicted so that maintenance and mitigating measures can be optimised.

In this article a basic rail break prediction model based on historic data is presented.

## 2 THE TFR ORE LINE

The Ore Line, also known as the Saldanha Sishen line, is the second longest heavy haul line in the world. It boasts 861 km of single track, with train crossing points (loops) at roughly 40 km intervals. This line carries the longest production trains worldwide at over 4 km in length and at 30 tonnes per axle load.

## 2.1 The Ore Line as a unique system

The Ore Line presents an assortment of engineering and maintenance challenges with its unique combination of continuous welded rail (CWR), high axle loading, multi-locomotive trains (Zhuan 2007), and long distances between loops. Furthermore, a large proportion of the line is inland; temperatures reach extremes which cause rail temperatures to fluctuate severely. The axial stress in the rail depends on the temperature of the rail, and therefore CWR, in contrast to rail with expansion joints. CWR experiences large changes in axial stress as temperatures fluctuate. Fluctuations in axial stress on the Ore Line have traditionally led to a large number of rail breaks (Duvel 2015).

Studies have shown that rail breaks are the most significant single contributor to derailments on the Ore Line (Duvel 2015). Furthermore, the probability of a rail break translating into a derailment on this line is high compared to other heavy haul lines (Duvel 2015). Since derailments on the Ore Line are a function of the incidence of rail breaks, focus has been placed on reducing the incidence of rail breaks as well as intelligent detection of rail breaks over the last decade.

## 2.2 Rail break formation

Heavy axle loads induce high stresses in the CWR (Zhao & Stirling 2006). Where existing defects (such as fractures and cracks) exist, stress concentrations are induced due to cyclic forces (caused by repeated train passage), wich can lead to rail breaks (Jeong 2003).

The stress free temperature (SFT) of a portion of rail is the temperature at which the rail was inserted and where the rail force is approximately zero (Gräbe et al. 2007). The difference in temperature between the current temperature of the rail and the SFT causes tensile or compressive forces in the rail.

A rail break usually occurs under tensile conditions (Gräbe et al. 2007) and therefore the two newly created ends of the rail are pulled apart. The further these are pulled apart, the more severe the break.

## 2.3 Rail break and derailment prevention on the Ore Line

To mitigate derailment risk due to rail breaks on the Ore Line, the following processes have been put in place. Some are done all year round and others only during specific, high rail break risk months:

1. Throughout the year, an ultrasonic measuring vehicle is run at least once a month to identify defects that pose a rail break risk – after each run the identified high-risk defects are removed.
2. Longitudinal strain measurement systems are installed at 1 km intervals along the track. These are designed to track changes in tensile forces in the rail and are used for maintenance purposes (when strain levels are above certain limits, de-stressing is performed). Unfortunately, these systems are not currently utilised optimally due to theft as well as constraints with regards to initializing the systems with the correct SFT of the rail.
3. A broken rail detection system, namely the Ultrasonic Broken Rail Detection (UBRD) version 4, has been installed throughout the Ore Line (at 1 km intervals). This system is also not currently used to its maximum potential due to theft of equipment, communication failures and maintenance challenges. Version 5 is expected to address some of these issues and also reduce the number of UBRD installations required.
4. In the winter months in the morning hours, trolleys are run ahead of each loaded train to identify rail breaks. This is a very manual and time consuming process. This process also potentially reduces the number of loaded trains that can be run weekly, which impacts freight volumes negatively.
5. Speed restrictions for loaded trains are imposed during the same time periods as when the trollies are run and also impact freight volumes. Lowering loaded train speeds by 20 km per hour reduces the forces on the rail (which slows defect growth and reduces rail break severity); it also reduces the impact of any derailments that may occur.

## 2.4 The need for a predictive system

Whereas the mitigating measures have reduced the annual number of rail breaks on the Ore Line over the last decade, it is believed that the number of rail breaks can be further reduced by optimising these measures based on a rail break risk model.

A fully predictive model (where trains will be stopped or slowed in high risk circumstances) is the most comprehensive solution to the rail break and rail break-caused derailment problem. The model presented in this article is the first step in this direction.

## 3 THEORETICAL CONCEPTS RELATING TO PROBABILISITC GRAPHICAL MODELING

An understanding of the following terms and concepts are required prior to understanding modelling technique that is presented in this article.

### 3.1 Probability theory concepts and definitions

The core probability theory definitions and concepts are now presented. They are referred to with regards to events $x$ and $y$, but these concepts and definitions can also be applied directly to the random variables (RVs) by replacing the events $x$ and $y$ with the random variables $X$ and $Y$. RVs can assume multiple states and the conditions must hold true for all combinations of states.

Conditional probability: The probability of an event *x* occurring while constrained by the occurrence of an event *y* is the conditional probability:

$$p(x \mid y) = \frac{p(x, y)}{p(y)} \quad (3.1.1)$$

where $p(x, y)$ is the joint distribution of events *x* and *y*.

Because $p(x, y) = p(y, x)$, (3.1.1) can be written as:

$$p(x \mid y) = \frac{p(y \mid x) p(x)}{p(y)} \quad (3.1.2)$$

Equation (3.1.2) is known as Bayes' rule. Another way of viewing Baye's rule is in terms of:

- the *prior probability* (the probability of our hypothesis, *H*, being true without knowing anything about our current evidence, *E*),
- *posterior probability* (the probability of *H*, given *E*),
- *likelihood* (the probability of the *E* given *H*) and the
- *marginal likelihood* (the probability of *E* – independent *H*).

This representation of Baye's rule can be seen in Equation 3.1.3.

$$p(H \mid E) = \frac{p(E \mid H) p(H)}{p(E)} \quad (3.1.3)$$

Marginalisation: Given the joint distribution $p(x, y)$,

$$p(x) = \sum_y p(x, y) \quad (3.1.4)$$

$p(x)$ is the marginal of $p(x, y)$. The same applies for $p(y)$: marginalize out *x* by summing out *x* over the joint distribution.

Independence and conditional independence: Equation 3.1.1 can be re-arranged to give $p(x, y) = p(x \mid y) p(y)$. If events *x* and *y* are independent, this becomes,

$$p(x, y) = p(x) p(y) \quad (3.1.5)$$

by the definition of independence.

In probabilistic graphical modeling, there are often cases where two events are dependent on each other, but are *conditionally* independent, given another event, *z*. If *x* is conditionally independent on *y* given *z* then:

$$p(x, y \mid z) = p(x \mid z) p(y \mid z) \quad (3.1.6)$$

In terms of RVs, we can interpret equation 3.1.6 in the following way: "knowing the state of random variable *Z*, we learn nothing more about the state of *X* if we know the state of *Y*".

## 3.2 *Graph theory concepts*

Probabilistic graphical modelling independence assumptions and causality to be represented in a graphical way using graph theory techniques. Some of the basic graph theory concepts required to understand the remainder of the article are covered in this section.

- A mathematical graph, $G(V, E)$, can be described by a set of vertices, $V = \{v_1, v_2, v_3, ..., v_n\}$, and edges, $E = \{e_1, e_2, e_3, ..., e_n\}$, where each edge connects two vertices (also known as nodes).
- If there is at most one edge between each pairs of vertices, the graph is singly connected.
- Edges can have a direction or no direction.
- A graph in which all edges are directed is called a directed graph.
- Path: A finite sequence consisting of alternating vertices and edges where no two edges or vertices are repeated (except possibly for the first and last vertex).
- Cycle: A directed path that starts and ends at the same vertex is called a cycle.
- Acyclic graph: A graph that has no cycles.
- Parent node: A parent node is a node that has a directed edge pointing from it to one or more other nodes.
- Child node: A node that has a directed edge pointing towards it from one or more parent nodes.

## 3.3 *Bayesian Networks*

Probabilistic Graphical Models (PGMs) can be seen as data structure for encoding probability distributions. A Bayesian Network, also called a Bayes net (BN) or Belief Network, is a specific type of PGM that can be represented by a directed, acyclic graph (DAG).

BNs allow one to represent independence assumptions in a probability distribution. For a distribution, *P*, these independence assumptions can be used to form an I-map (independency map), *G*, for *P*, if *P* is representable as a BN over *G*. We represent the BN, *G*, as a DAG from which we can read off the conditional dependence assumptions directly.

In a BN, each node represents a variable (along with its conditional dependence assumptions) and the directed edges represent direct dependence between parent and child nodes. In the example BN shown Figure 3.3.1 (a), we see nodes $x_1$ and $x_2$ are parents of $x_3$. This means that $x_3$ depends directly on $x_1$ and $x_2$. Both $x_1$ and $x_2$ have no parents themselves – this means that they are independent of any other variables in the network (given that we know nothing about their child or child's child, $x_4$).

The probability distribution of a Bayse net, *p(x)*, can be described by the following equation:

$$p(x_1,...,x_D) = \sum_{i=1}^{D} p(x_i \mid pa(x_i)) \qquad (3.3.1)$$

Where *pa(x$_i$)* represents the parent variables of *x$_i$*. This equation allows us to be able to read off the distribution of any Bayse net by looking at the structure of the graph.

In Figure 3.3.1 (a) we see the graphical representation of a BN and in (b) we see the mathematical representation of the distribution for (a) as derived from Equation 3.3.1.

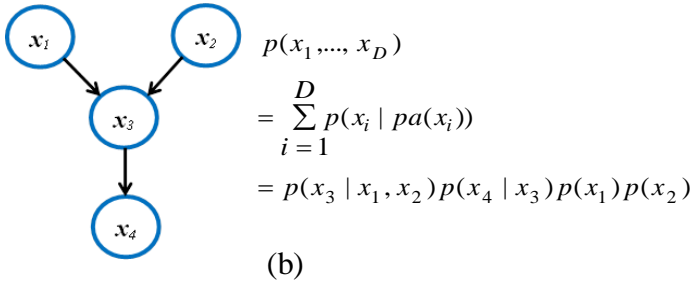

(a)

(b)

$$p(x_1,...,x_D)$$
$$= \sum_{i=1}^{D} p(x_i \mid pa(x_i))$$
$$= p(x_3 \mid x_1, x_2)p(x_4 \mid x_3)p(x_1)p(x_2)$$

Figure 3.3.1: (a) A belief network represented by the distribution shown in (b).

## 4 CREATION OF THE MODEL

The challenge in creating any PGM or BN is correctly using the available data as well as expert knowledge to construct an appropriate model. Rail break formation on the Ore Line is part of a complicated system that has causal relationships between RVs. In this section we list some of these RVs with their dependencies and the present the challenges involved in creating a BN to model rail break risk prediction.

### 4.1 *Dependencies and causality*

To start the modelling process the question asked was: "What factors or variables cause rail breaks to occur?". To answer this question both literature and rail engineering experts were consulted.

The random variables *Strain* (*St*), *Change in Strain* (*ΔSt*), *Rate of Change in Strain* (*rΔSt*), *Temperature* (*Tp*), *Change in Temperature* (*ΔTp*), *Rate of Change in Temperature* (*rΔTp*), *Season* (*S*), *Time of Day* (*T*), *Location* (*L*), *Gross Tonnage Moved* (*GT*), *Rail Age* (*A*), *Maintenance Activities* (*M*), *Defects* (*D*) and *Rail Type* (*W*) were identified. It is evident that most of these variables depend on one another. Some of the relationships are highly complex and also cyclical in nature. This makes modelling all of these dependencies in a BN almost impossible.

In Table 4.1.1 each variable is listed and the believed causality between the variables are defined by listing possible parent (*pa*) and child (*ch*) nodes for each variable. Information about these RVs and their influence on the occurrence of rail breaks is not available for each of the RVs. In the table below we summarise the dependencies and indicate if the necessary information regarding these variables is currently available (*A*). For simplicity, we summarise *St*, *ΔSt*, and *rΔSp*, as *St* and *Tp*, *ΔTp* and *rΔTp* as *Tp*.

Table 4.1.1: Table showing the critical RVs, the parent / child dependencies between RVs and the data availability for each RV.

| RV | Explanation | Pa | Ch | A |
|---|---|---|---|---|
| St | *Strain:* Tensile forces in the rail make it more likely for an existing defect or crack in the rail to become a rail break. | Tp, M | D, R | N |
| Tp, | *Temperature:* Ambient temperature. | T, S, L | St | N |
| T | *Time of Day:* Certain periods of the day see more rail breaks than others. | | Tp, | Y |
| S | *Season:* More breaks seen in certain months that we can group into seasons. | | Tp, | Y |
| L | *Location:* (1) Coastal areas are less affected Temperature fluctuations than in-land regions. (2) The layout of the Ore Line is such that there are steeper gradients and sharper turns in the rail in the inland portion of the line than the coastal section. | | (1) Tp, (2) R, D | Y |
| GT | *Gross Tonnage:* The entire line sees similar tonnages moved annually. Due to scheduled rail replacement, however, not all rail is the same age and therefore certain parts of the rail have experienced a greater gross tonnage. | A, M | D, R | Y |
| M | *Maintenance:* Interventions such as rail replacement, tamping, grinding. The interaction between maintenance and the other RVs is complex and involves feedback. | D, R, L, S, A | D, R, GT | N |
| D | *Defects:* Existing ultrasonic defects and geometric deviations as well as the history of these defects and deviations (trends). | GT, A, M, St, L | M, R | N |
| W | *Rail Type (welds):* On the Ore Line, only 60kg/m is used. Parts of the rail are welds (either alluthumic or flash-butt). | | D, M, R | N |
| R | *Rail Break:* All of the above RVs influence the probability of a rail break occurring, but most of them, indirectly. | D,G T, St, W | M | |

In Table 4.1.1 we make assumptions when defining our proposed causality structure, one of which is that temperature, change in temperature and rate of change in temperature affect the strain in the rail and cause defects to occur but do not directly cause rail breaks. We assume that strain as well as change in strain and existing defects, however, can directly result in rail breaks.

Other assumptions include the assumption that maintenance activities affect the probability of a rail break occurring and that the number of detected defects in a section is correlated with the likelihood of a rail break occurring in the same section.

### 4.2  Selecting RVs to use in the basic model

Once causalities and dependencies have been assigned and relevant assumptions are made to simplify the model, we need to evaluate the available data and collected expert knowledge.

The available data that was used is historic rail break data on TFR's SAP system from April 2006 to end March 2016 (10 financial years). The expert knowledge used is information about section lengths, rail break mitigating measures and the train schedule.

For the Ore Line, we do not currently have sufficient data available in the SAP database for many of the RVs in Table 4.1.1. For some of the RVs, we have some, but not all of the data necessary to incorporate the RV directly into our model. For example, we have information as to how many rail breaks occurred in Thermit welds, Flash butt welds and Parent rail, (the *likelihood*), but we do not know exactly how many of each of these types of rail there are along the line (the *marginal likelihood*) . We could estimate these values, but this is not in the scope of this model.

An alternative method to estimating unknown values is to include these un-observed RVs (latent variables) in the model by *learning* the appropriate model structure from the data. For this article, we omit the latent variables such as rail temperature and strain, but in future work they will be included.

In our model, to include a RV, we require:
- The *marginal likelihood*, $p(E)$, of a RV: e.g. the probability of it being a certain season.
- The *posterior probability*, $p(H|E)$ with respect to the *hypothesis*, Rail Break: e.g. the probability that a rail break will occur, given that it is a certain season. If the *posterior* is not available, and we have the *likelihood* and the *prior,* we can use Bayes' rule to calculate the *posterior*.

Even when we have the *marginal likelihood* as well as the *posterior probability* for an individual RV with respect to *Rail Break*, we are still not guaranteed an accurate model. We are less interested in the effect of the season or time of day on the probability of a rail break occurring than in how the combined effect of these RVs on the posterior probability. In cases where the joint distributions are unknown, one can make independence assumptions and model the RVs separately.

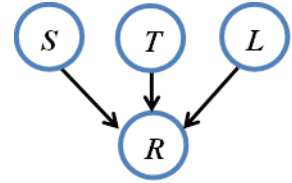

Figure 4.2.1: (a) A 4-node BN with *Season, Time of Day* and *Location* as parents of node *Rail Break*.

For our model, however, we have all the required data available to construct the model directly from the data for our chosen RVs. The chosen RVs are shown in Figure 4.2.1 and the values that the RVs can assume are shown in Table 4.2.1. The joint distribution for Figure 4.2.1 derived using Equation 3.3.1 is:

$$p(S,T,L,R) = p(R \mid S,T,L)p(S)p(T)p(L) \quad (4.2.1)$$

By chosing this structure (Fig. 4.2.1) we imply that *Season*, *Time of day* and *Location* are independent of one another and that *Rail Break* depends on all three of these RVs.

Table 4.2.1: Table showing the chosen RVs for our 4-node BN as well as the values that the RVs can assume.

| RV | Name | Values that the RV can assume |
|---|---|---|
| S | Season | Early summer ($S_0$), late summer ($S_1$), winter ($S_2$), late winter ($S_3$) |
| T | Time of day | Morning ($T_0$), remainder of the day ($T_1$) |
| L | Location | Coastal ($L_0$), semi-coastal ($L_1$), inland ($L_2$) |
| R | Rail break | No broken rail ($R_0$), broken rail ($R_1$) |

### 4.3  Choosing $p(R)$

Up to this point we have not clearly defined the precise meaning of $p(R)$ (the probability of a rail break occurring). In defining $p(R)$ we are confronted with two questions:
1. What is the time-frame in which we are interested in the rail break occurring?
2. Are we interested in the probability of a rail break occurring in the entire line, a portion of the line (such as a loop or group of loops) or a small section of the line (say the size of a weld / defect)?

The answers to the above two questions depend on how the model is to be utilised. If we are interested in knowing the chances of a rail break occurring between maintenance activities such as de-stressing or monitoring activities such as running the ultrasonic measuring car, we would be interested in bi-annual or monthly intervals. If we wanted a live system, we could be interested in the probability of a rail breaking before or under the next loaded train.

Similarly, location can be chosen as an entire section, loop or the distance between two ultrasonic broken rail detection installations (about 1 km apart).

In future work, where UMC data and broken rail data are included in the model, we may choose $p(R)$ as the probability of a rail break occurring before or under the next loaded train within a UBRD section. In this model, however, we are interested in the probability of a rail break occurring before or under the next loaded train, but relax the location criteria to be one of three sections large sections (Table 4.2.1).

We use historic rail break data as well as an estimate of the number of trains traveling daily to calculate $p(R)$. We know how many of the trains over the 10 year period experienced a rail break (or where there was a break detected between the previous train and the train of interest) and how many trains did not encounter a break (or were delayed because of a break). To calculate $p(R)$, we use the proportion of trains that can be linked to a rail break compared to those that cannot.

### 4.4 Choosing Location, Season and Time of Day

To investigate possible values for *Location*, *Season* and *Time of Day*, SAP data was used to plot graphs grouping the number of rail breaks into different buckets and then deciding on appropriate groupings for each RV. We briefly discuss how each of these groupings were chosen.

**Season:** By looking at the available rail break data (shown in Figure 4.4.1), we see that most rail breaks occurred between April and July and that the fewwest occurred in between October and December. Based on the average number of monthly breaks, we divided the year into four seasons (not of equal lengths). If the defined seasons were of equal lengths, the *marginal likelihood* (or *evidence*) would have had a flat distribution of ¼ for each season. In our case, our distribution is not flat but based on the number of trains that were moved in each of these seasons.

It is important to note that this rail break data gives us the probability of it being a certain season, given that we know that a rail break has occurred and not the probability of a rail break actually occurring in a specific season.

**Time of Day:** A histogram was plotted showing the average number of breaks for each hour bucket. There was a clear discrepancy between the morning hours between 4am and 11am and the rest of the day. Table 4.4.1 shows the percentage of breaks in each of the two groupings as well as the proportion of the day belonging to each grouping (from which the *marginal likelihood* can be calculated). The normalized percentage shows how much more likely rail breaks are to occur in the morning than in the rest of the day.

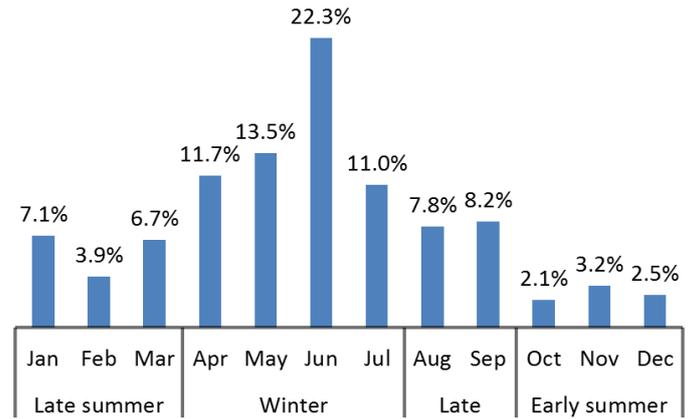

Figure 4.4.1: Figure showing the percentage of annual rail breaks per month, as well as the season allocations, for the last 10 financial years.

Table 4.4.1: Table showing the two *Time of Day* groupings. The percentage of rail breaks the percentage of the day and the normalized percentage is shown for both groupings.

| Time of day | % of rail breaks | % of day | Normalized percentage |
|---|---|---|---|
| *Morning (4am - 11am)* | 56% | 29% | 76% |
| *Not morning (11am - 4am)* | 44% | 71% | 24% |

**Location:** The Ore Line is often grouped into the section from Salkor yard to Loop 3 (Bamboesbaai), Loop 3 to Loop 10 (Halfweg) and Loop 10 to Erts. The shape of the Ore Line is such that the first section of the line follows the coast. For the second section and third sections, the line heads inland towards the mines. In the data set used, considerably more rail breaks occurred in the most inland section. This is due, partially, to the extreme temperatures seen in this section and partly due to the line layout (curves and steep gradients) for this portion of the line [3].

### 4.5 Computing the joint distribution

Now that the RVs have been defined, the joint distribution can be calculated based on these RVs.

For each train that experienced a break we have information as to in which section the break occurred ($L$), the time of the day ($T$) and the season ($S$) in which the break occurred. For every train that did not encounter a break we also have values for $T$ and $S$. To calculate the joint distribution for each section, we create a table with each combination of possible RV values and the number of trains that can be assigned to each combination. Normalizing this table gives us the joint distribution over the RVs.

The joint distribution was read into PGM software, EMDW, so as to compute the marginal and conditional distributions required to extract information from the model.

## 5 RESULTS

The BN was used to evaluate rail break risk, firstly for the general case where we know nothing about the states of the RVs and then for specific cases.

### 5.1 General rail break risk

To interpret the results from the BN, we need to keep our definition of $p(R)$ in mind. Our model predicts the probability of a specific train encountering a rail break. Practically that means that each train is exposed to the risk of a rail break occurring under it or before it enters the section – this is the risk we would like to predict.

If we know nothing about the state of the RVs, we can make a prediction about this risk by marginalising out all the RVs except $R$. The distribution we are left with after marginalisation is $p(R)$. Using EMDW to query the BN, we find that $p(R_1)$ (the probability of a rail break occurring for a specific train) is 1.9%. Changing the values of the RVs change this probability and give us an indication of the risk for different scenarios.

### 5.2 Rail break risk for specific cases

Because each train traverses the entire line (in the general case), the probability that a rail break occurs for a specific train anywhere in the entire line is the sum of the probabilities of a rail break occurring in each of the three sections.

We now look primarily at only the inland, high risk section of the Ore Line, namely between loop 10 and Erts. The probability of a break occurring in this section is 10 times more likely than for the coastal section of the line (given that we know nothing else about the conditions at the time). We now list some of the noteworthy findings:

1. A train going through the inland region has 2.4% chance of experiencing a rail break in winter and 1.4 % in late winter as opposed to 0.3% chance in early summer.
2. If a train makes its whole trip through the coastal section in non-morning hours, it only has a 0.7% chance of rail break occurring as opposed to a 3% chance of experiencing a rail break if it spends the entire morning in the inland section.
3. A train traveling through inland portion of the line in winter, has a 5,4% chance of experiencing a rail break if it spends its entire morning in the inland section. This can be contrasted with the 0.07% probability of a rail break occurring in the case where a train makes its whole trip through the coastal section in non-morning hours in the early summer.

## 6 CONCLUSION

In this article, the factors that affect rail breaks and potentially derailments were discussed and the viability of including these factors into a rail break prediction model was investigated. A basic BN incorporating some of these factors was developed by using historic rail break data.

The model was used to predict the rail break risk that each train experiences under varying circumstances. This modelling technique allows us to clearly see that specific locations, times of the day and seasons increase the rail break and therefore derailment risk significantly.

By predicting the rail break risk that each train introduces under a combination of circumstances, we can ultimately implement mitigating measures (such as speed restrictions) more efficiently than by analysing the individual influences of factors in insolation.

There is currently work being done to integrate more of the factors discussed in this article to create a more comprehensive, flexible model so as to make more detailed predictions.